\documentclass[preprint,12pt]{elsarticle}
\pdfoutput=1

\usepackage{amssymb}
\usepackage[ruled,linesnumbered]{algorithm2e}
\usepackage{booktabs}
\usepackage{subfig}
\usepackage{tabulary,graphicx,times,caption,fancyhdr,amsfonts,amssymb,amsbsy,latexsym,amsmath}
\usepackage[utf8]{inputenc}
\usepackage{url,multirow,morefloats,floatflt,cancel,tfrupee,textcomp,colortbl,xcolor,pifont}
\usepackage[nointegrals]{wasysym}

\begin{document}

\begin{frontmatter}



\title{A Simple Geometric-Aware Indoor Positioning Interpolation Algorithm Based on Manifold Learning}


\author[1,2]{Suorong Yang}
\author[1,3]{Geng Zhang}
\author[4]{Jian Zhao}
\author[1,3]{\\ Furao Shen \corref{cor1} }

\address[1]{State Key Laboratory for Novel Software Technology, Nanjing University, China}
\address[2]{Department of Computer Science and Technology, Nanjing University, China}
\address[3]{School of Artificial Intelligence, Nanjing University, China}
\address[4]{School of Electronic Science and Engineering, Nanjing University, China} 
\cortext[cor1]{Corresponding author.  E-mail address: frshen@nju.edu.cn (F. Shen).}
  \emailauthor{sryang@smail.nju.edu.cn}{S. Yang}
  \emailauthor{zhanggeng@smail.nju.edu.cn}{G. Zhang} 
  \emailauthor{jianzhao@nju.edu.cn}{J. Zhao}
  \emailauthor{frshen@nju.edu.cn}{F. Shen}

\begin{abstract}
Interpolation methodologies have been widely used within the domain of indoor positioning systems.
However, existing indoor positioning interpolation algorithms exhibit several inherent limitations, including reliance on complex mathematical models, limited flexibility, and relatively low precision.
To enhance the accuracy and efficiency of indoor positioning interpolation techniques, this paper proposes a simple yet powerful geometric-aware interpolation algorithm for indoor positioning tasks.
The key to our algorithm is to exploit the geometric attributes of the local topological manifold using manifold learning principles.
Therefore, instead of constructing complicated mathematical models, the proposed algorithm facilitates the more precise and efficient estimation of points grounded in the local topological manifold.
Moreover, our proposed method can be effortlessly integrated into any indoor positioning system, thereby bolstering its adaptability.
Through a systematic array of experiments and comprehensive performance analyses conducted on both simulated and real-world datasets, we demonstrate that the proposed algorithm consistently outperforms the most commonly used and representative interpolation approaches regarding interpolation accuracy and efficiency.
 Furthermore, the experimental results also underscore the substantial practical utility of our method and its potential applicability in real-time indoor positioning scenarios.
\end{abstract}




\end{frontmatter}
\section{Introduction}
Indoor positioning has garnered significant attention within the realm of intelligent Internet of Things (IoT) applications~\cite{iot}, facilitating developments in areas such as intelligent robotics,  factory automation, autonomous driving, and pedestrian tracking~\cite{indoor1,indoor2,LAN2023126545,zafari2019survey}.
However, while outdoor environments benefit from the Global Positioning System (GPS) positioning technologies, indoor spaces present a unique set of challenges for positioning due to the absence of a direct line of sight and the presence of obstacles such as walls~\cite{LAN2023126545}, which diminishes the effectiveness of GPS systems.
Consequently, there has been a growing reliance on indoor positioning systems (IPSs).
Despite their increasing prevalence and diverse applications, IPSs are still vulnerable to wireless signal disruptions caused by architectural characteristics and environmental dynamics~\cite{influence,influence2}.
These disruptions encompass phenomena such as multipath effects~\cite{multipath}, non-line-of-sight transmission~\cite{NLOS2}, and device heterogeneity~\cite{heterogenity}, which impose significant constraints on the advancement of indoor positioning techniques~\cite{ipf,ipf2,ipf3}.
For instance, for real-time miner positioning in intelligent underground mining scenarios, the interference of wireless signals critically jeopardizes safety protocols, leading to the loss of real-time tracking of miners and posing severe risks to operational safety.
Addressing these disruption factors is paramount to furthering the development and ensuring the reliability of indoor positioning techniques in intelligent IoT applications.

Therefore, interpolation algorithms emerge as promising solutions for estimating unobserved signal values and have gained widespread adoption in various IPSs~\cite{app,app2}.
Nevertheless, it is imperative to recognize the inherent limitations associated with existing interpolation methods. 
Firstly, most existing interpolation techniques rely on intricate mathematical functions while disregarding the intrinsic local geometric characteristics of the positioning trajectory~\cite{makima,cubic_spline}. 
Consequently, these methods entail extensive equation solving, incurring significant computational overhead and limiting their applicability within indoor smart IoT applications.
For instance, cubic spline interpolation~\cite{cubic_spline} and modified Akima interpolation (Makima)~\cite{makima} construct trajectory models for interpolating missing points using high-order polynomial functions, which do not account for the physical characteristics of the moving object, resulting in relatively low interpolation accuracy and efficiency.
Secondly, the exploration of interpolation methods in indoor positioning based on machine learning techniques has been relatively scant and nearly absent in the existing literature.
Furthermore, previous interpolation methods typically adopt distinct strategies based on whether the missing point falls within or outside the observation range, diminishing their overall flexibility.

To address these limitations, we propose a simple yet powerful geometric-aware interpolation algorithm tailored specifically for indoor positioning scenarios.
Based on manifold learning principles within the realm of machine learning~\cite{LLE}, the proposed algorithm learns the local topological manifold of positioning trajectories.
By recovering the global nonlinear structure through the local linear fitting, our algorithm facilitates precise interpolation of missing data points within the manifold, obviating the need for extensive equation solving. 
Consequently, the proposed algorithm offers enhanced computational efficiency, making it well-suited for real-time indoor positioning systems deployment.
Meanwhile, our method utilizes a uniform strategy to address missing points, irrespective of their location in relation to the observation range, enhancing its flexibility. 
Moreover, our algorithm is also agnostic to the indoor positioning framework types, relying solely on positioning data points, thereby ensuring seamless integration into any indoor positioning framework (e.g., Wi-Fi-based IPSs) and conferring a high degree of flexibility.
To evaluate the effectiveness of our algorithm, we conduct extensive experiments using both simulated and real-scene data. 
The experimental results demonstrate that our algorithm significantly outperforms other most commonly used and representative interpolation methods, notably in accuracy and computational efficiency. 
These findings highlight the potential practical applications of our method, thereby contributing to artificial intelligence-driven IoT applications.
To summarize, we highlight the key contributions of this paper as follows:
 \begin{itemize}
 	\item  We propose a new geometric-aware interpolation algorithm for indoor positioning that learns the local topological manifold of the positioning trajectory, enhancing both interpolation accuracy and efficiency.
 	\item The proposed algorithm demonstrates enhanced flexibility by applying equally to estimated points, whether within or outside the range of historical data, but also possesses the versatility to be seamlessly integrated into any IPS.
 	\item We systematically compare our algorithm with the most prevalent and representative interpolation methods, affirmatively demonstrating that our approach excels in performance.
  Specifically, it showcases superior accuracy, stability, and computational efficiency compared to other methods.
 \end{itemize}

 This paper is organized as follows. Section~\ref{sec:related-work} introduces the related work. Section~\ref{sec:LLI} presents the proposed algorithm in detail.
 Section~\ref{sec:experiment} describes the experimental settings, and the experimental results are comprehensively analyzed and compared.
 Section~\ref{discuss} discusses the challenges and potential future works.
 Conclusions are drawn in the last section.
\section{Related Work}\label{sec:related-work}
This section presents an overview of the most representative and commonly used interpolation methods within the realm of indoor positioning tasks. 

Interpolation is a technique that can solve the curve passing through some known knots and predict the function values of unknown positions according to the curve obtained~\cite{survey_inter}.
The fundamental principle underlying interpolation is to infer data points based on some provided ones.
In indoor positioning, the k-Nearest Neighbors (KNN) interpolation algorithm~\cite{NN1} involves predicting a location based on the $k$ latest known data points.
 Linear interpolation~\cite{linear_interpolation1} estimates the intermediate point along a straight line formed by its front and back points, assuming a constant gradient in the rate of change between the two points. 
 However, linear interpolation is inadequate for complex and random trajectories, which are typical in indoor positioning. 
 Kriging interpolation~\cite{kriging} calculates the weighted average of the known sample values around the point of interest to generate an optimal unbiased estimate. 
By minimizing the error variance and setting the average of the prediction errors to zero, Kriging interpolation can obtain the weight for optimal unbiased estimates.
However, Kriging interpolation assumes that any point $(x,y)$ in space has the same expectation and variance, which is not always valid in indoor positioning where the trajectories are locally correlated. 
Radial basis function (RBF) interpolation combines several deterministic functions~\cite{rbf,rbf2}.
Given a set of $n$ positioning points $\left\{x_{i}\right\}_{i=1}^{n}$ and corresponding values $\left\{y_{i}\right\}_{i=1}^{n}$, RBF interpolation is:
\begin{equation}
	s(x)=\sum_{i=1}^{n} \lambda_{i} \phi\left(\left\| x-x_{i}\right\|\right)
\end{equation}
where $ \phi(r), r>0$ is some radial function, and coefficients $\lambda_{i}$ is determined by $s(x_i)=y_i$.
Compared with linear interpolation and Kriging interpolation, RBF interpolation can obtain better performance.
 However, the accuracy of an RBF method is highly sensitive to the choice of basis functions and shape parameters~\cite{rbf_basis}, which drastically limits its flexibility.
Cubic spline interpolation~\cite{cubic_spline,szabo2021closed} uses a series of unique cubic polynomials to fit the curves between each data point.
The continuity of each polynomial's first and second derivatives guarantees the generated curve's continuity and smoothness. 
Therefore, compared with RBF interpolation, the recorded data can automatically determine the form of piecewise functions, while the continuity of the interpolation function comes with high computation costs. 
To balance the continuity of interpolation function and calculation costs, piecewise cubic Hermite interpolation polynomial (PCHIP)~\cite{hermite} was proposed.
PCHIP also uses piecewise cubic polynomials to generate functions between every two consecutive points, but only the first derivative is restricted to be continuous.
In addition, the Hermite formula is applied to each interval to ensure the smoothness of the generated curve, which means that the curve and its first derivative are continuous.
Makima~\cite{makima} is an improved version of Akima interpolation~\cite{Akima}. 
Akima interpolation uses a unique cubic polynomial $P(x)$ to determine the spline curve between every pair of consecutive points. 
The Akima interpolation algorithm can balance the spline interpolation and PCHIP interpolation methods.
The fluctuation of Akima interpolation is smaller than the amplitude of the spline interpolation, and Akima interpolation is not as aggressive as PCHIP in ensuring continuity~\cite{Akima}.
Specifically, the coefficients of polynomials are determined by four constraints: $P(x_i)=y_i, P(x_{i+1})=y_{i+1}, P'(x_i)=s_i,$ and $ P'(x_{i+1})=s_{i+1}$, where $s_i$ is defined as the following weighted average of $m_{i-1}$ and $m_i$:
\begin{equation}
	\begin{aligned}\label{akima}
		s_i=\frac{|m_{i+1}-m_i|m_{i-1}+|m_{i-1}-m_{i-2}|m_i}{|m_{i+1}-m_i|+|m_{i-1}-m_{i-2}|}
	\end{aligned}
\end{equation}
where $m_i$ is the slope of the line segment from ($x_i,y_i$) to ($x_{i+1}$,$y_{i+1}$), namely $\frac{y_{i+1}-y_i}{x_{i+1}-x_i}$.
Under these conditions, the Akima spline function and its first derivatives are continuous. 
However, in some marginal cases, such as $m_{i+1}=m_i$ and $m_{i-1}=m_{i-2}$. $s_i$ in Equation~\eqref{akima} can not be calculated. 
To address the issue, Makima was proposed, which modifies the Akima's derivative formula by tweaking the weight $w_1$ and $w_2$ of the slopes $m_{i-1}$ and $m_{i}$:
\begin{equation}\label{akima2}
	\begin{aligned}
		s_i=\frac{w_1}{w_1+w_2}m_{i-1}+\frac{w_2}{w_1+w_2}m_{i}
	\end{aligned}
\end{equation}
where $w_1=|m_{i+1}-m_i|+\frac{|m_{i+1}+m_i|}{2}$ and $w_2=|m_{i-1}-m_{i-2}|+\frac{|m_{i-1}+m_{i-2}|}{2}$. 
Makima spline only uses the values from neighboring points to construct the coefficients of the interpolation polynomial between any two points, which means that there is no extensive system of equations to be solved.
In addition, the Makima spline avoids non-physical fluctuation in regions where the second derivative of the basic curve changes rapidly.

Neural network approaches interpolate points by learning spatial patterns from wireless signal strengths to predict locations.
Nevertheless, they are not commonly used for indoor positioning interpolation due to concerns like overfitting, dynamic indoor environments, computational overhead, and latency.
For instance, a model trained on data from one time may not perform well later unless continuously retrained.
Similarly, models trained for a particular scenario may exhibit suboptimal performance when confronted with dissimilar scenarios unless the environmental conditions closely resemble one another.
Instead, we propose a manifold learning-based interpolation algorithm, which dynamically learns the current trajectory manifold and interpolates new data points.
In summary, existing techniques either require an extensive system of equations to solve or show insufficient performance for indoor positioning interpolation. In this paper, we propose a novel manifold learning-based algorithm to avoid the disadvantages mentioned above.

\section{The Proposed Algorithm}\label{sec:LLI}
\begin{figure}[h]
	\centering
	\includegraphics[width=0.6\textwidth]{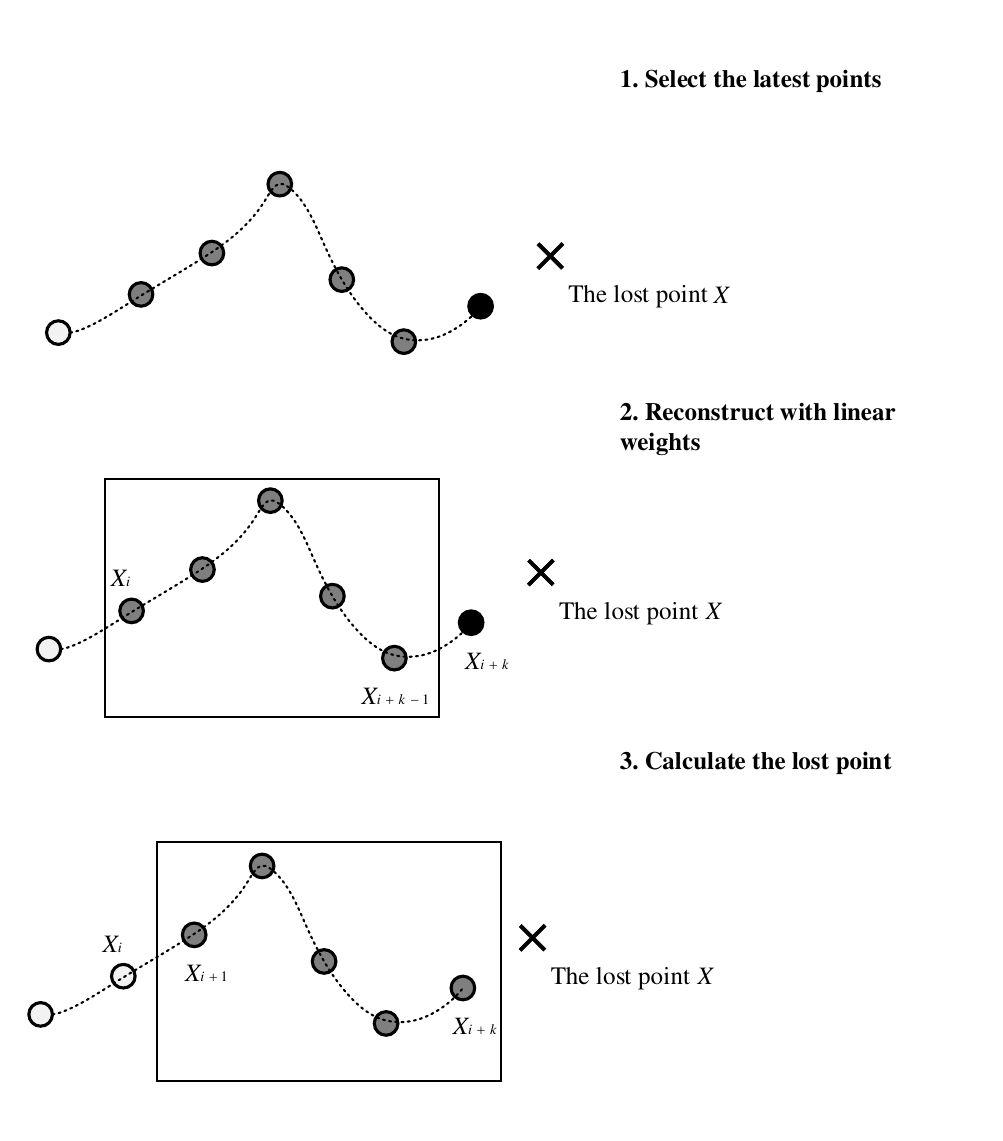}
	\caption{The general workflow of the proposed algorithm: (1) selecting previous $k$  points before the missing point $X$. (2) Calculate the optimal adaptive weights of linear reconstruction $X_{i+k}$ from its neighbors in the time domain. (3) Calculate the coordinates of the missing point $X$ using the adaptive linear coefficients. The square is sliding forward by one point and uses the points inside it to calculate the lost point $X$.}
	\label{LLI}
\end{figure}
In this section, we present the simple yet powerful geometric-aware interpolation algorithm for indoor positioning tasks.
Summarized in Figure~\ref{LLI}, the proposed algorithm estimates missing or unknown data points based on the geometric patterns of the latest historical data. 
Our algorithm is based on a basic assumption that the data points on the positioning trajectory lie in a local manifold constructed by neighbor points.
Therefore, the local manifold can be approximately linearly fitted~\cite{manifold,HAN2022877} and the missing points can be estimated based on the local manifold.
Finally, the global nonlinear structure can be recovered by the local linear fitting.
 
Considering a simple but representative indoor positioning scenario~\cite{2D}, suppose that the current trajectory data points $P$ consist of $N$ real-valued vectors, represented as a two-dimensional coordinate: $P=[\mathbf{p}_1,\mathbf{p}_2,...,\mathbf{p}_N]\label{eq1}$, where $\mathbf{p}_i$ is the location point of the trajectory at time $i$. 
Suppose that $\mathbf{p}_{N+1}$ is the missing point. Our work aims to estimate $\mathbf{p}_{N+1}$ based on the trajectory data points.
To maintain the continuity of the real-time trajectory, based on $\mathbf{p}_{N}$, we construct the lost point using its previous $k$ reference points in $P$:
\begin{equation} \label{eq2}
	\mathbf{p}_{N}=\sum_{i=N-k}^{N-1} w_{i}*\mathbf{p} _{i}
\end{equation}
where $w_i$ is the corresponding weight coefficient.
The coefficients represent the intrinsic geometric characteristics of each neighborhood~\cite{LLE}, which will be used to estimate the missing point.
The number of reference points, denoted as $k$, balances the time latency and the smoothness of the trajectory after interpolation, with a larger value of $k$ resulting in a smoother trajectory but with higher time latency, and vice versa.  
To construct a linear relationship between $\mathbf{p}_{N}$ and $k$ location points before it, we calculate the adaptive linear coefficients $W=[w_{N-k},w_{N-k+1},...,w_{N-1}]\in R^{k \times 1}$ of the Equation~\eqref{eq2}. 

Since the abscissa and ordinate of the positioning trajectory are independent, the proposed algorithm employs separate calculations for the abscissa and ordinate values.
We elucidate the method for determining the abscissa coefficients and mirror this exposition for the ordinate coefficients.
The optimization problem is formulated as follows:
\begin{equation}\label{opt_problem}
	\begin{aligned}
		 \min_{W} &||x_{i+k}- \sum_{j=0}^{k-1} w_{i+j}x_{i+j} ||^2 \\  
		 \text{s.t.}&\sum_{j=0}^{k-1} w_{i+j}=1, i=1,2,3,...,N-k.
	\end{aligned}
\end{equation}
where $x_{i+k}$ is the abscissa coordinate of the previous point to the missing data point, with a dimension of 1.
$\epsilon=||x_{i+k}- \sum_{j=0}^{k-1} w_{i+j}x_{i+j} ||^2$ is the reconstruction error function.
Thus, the essence of the reconstruction error resides in the square of the discrepancy between the lost point $x_{i+k}$ and the weighted summation of the abscissa values associated with the $k$ reference points.
Furthermore, as suggested in~\cite{LLE}, we rigorously constrain the least-square problem using an optimization constraint: $\sum_{j=0}^{k-1} w_{i+j}=1$. 
Finally, the optimal weights $W=[w_{i},w_{i+1},...,w_{i+k-1}]$ are found by solving the least-square problem ~\eqref{opt_problem}.

Specifically, we can calculate the adaptive linear coefficients by optimizing the regularized least squares objective:
\begin{equation}
	\begin{aligned}
		\epsilon&=||x_{i+k}- \sum_{j=0}^{k-1}w_{i+j}x_{i+j}||^2\\
		&=||X^\intercal  W||^2
	\end{aligned}
\end{equation}
where $W$ is $k$ abscissa coefficients of previous recorded points and $X=[x_{i+k}-x_{i},x_{i+k}-x_{i+1},...,x_{i+k}-x_{i+k-1}] \in R^{k \times 1}$. Then, we carefully construct the Lagrange function to calculate the coefficient vector reliably:

\begin{equation}\label{L_function}
	\begin{aligned}
		\mathcal{L}&=||x_{i+k}- \sum_{j=0}^{k-1}w_{i+j}x_{i+j}||^2-\lambda (\sum_{j=0}^{k-1}w_{i+j}-1)\\
		&=W^\intercal XX^\intercal W-\lambda \boldsymbol{1} W+\lambda
	\end{aligned}
\end{equation}
where $\boldsymbol{1}$ is all 1 vector of length $k$. 
To minimize the $\mathcal{L}$, we need to solve:

\begin{equation}\label{partial_derivative}
	\begin{aligned}
		\frac{\partial \mathcal{L}}{\partial W}&=XX^\intercal W-\lambda \boldsymbol{1}^\intercal \\
		&=CW-\lambda \boldsymbol{1}^\intercal  =0
	\end{aligned}
\end{equation}
where $C=XX^\intercal $. Based on Equation~\eqref{partial_derivative}, we obtain :

\begin{equation}\label{inver_C}
	W=\lambda C^{-1} \boldsymbol{1}^\intercal  \text{.}
\end{equation}
We know that $\sum_{j=0}^{k-1}w_{i+j}=\boldsymbol{1} W=1$. 
Therefore, we have

\begin{equation}
	\boldsymbol{1} \lambda C^{-1} \boldsymbol{1}^\intercal =1 \text{.}
\end{equation}
So,

\begin{equation}
	\lambda =(\boldsymbol{1}C^{-1}\boldsymbol{1}^\intercal )^{-1} \text{.}
\end{equation}
Finally,
\begin{equation}\label{coeffcients}
	\begin{aligned}
		W=\frac{C^{-1}\boldsymbol{1}^\intercal }{\boldsymbol{1}C^{-1}\boldsymbol{1}^\intercal }
	\end{aligned}
\end{equation}
by which we can accurately and efficiently determine the abscissa's adaptive weight vector. 
Similarly, the weight vector of the ordinate can also be obtained.
In this way, the local geometric structure adaptively determines the weight vector.
 It should be noted that the above derivation is based on the assumption that the matrix $C$ in Equation~\eqref{inver_C} is invertible. 
Here, we denote the dimension of each abscissa as $d$, and in our method, $d$ equals 1. 
However, if $k > d$, matrix $C$ may not be invertible.
To ensure reliable and efficient coefficient calculations, we include the trace of matrix $C$ as regularization: 
\begin{align}
	C=C+\sigma *\mathrm{trace}(C)
\end{align}
where $\sigma$ is used to control the regularization item.

Using the adaptive weighted coefficients $W_x$ and $W_y$, we can reconstruct the lost point while preserving the local topological manifold of the trajectory:
\begin{equation}
	\begin{aligned}
		&x=\vec{X_{i+1}}*W_x^\intercal \\
            &y=\vec{Y_{i+1}}*W_y^\intercal \\
	\end{aligned}
\end{equation}
where $\vec{X_{i+1}}$ is the abscissa vector starting from $x_{i+1}$ of length $k$, and  $\vec{Y_{i+1}}$ is the ordinate vector starting from $y_{i+1}$ of length $k$. 
Thus, our method creates a neighborhood-preserving map based on the derivation above.

\paragraph{Time Complexity Analysis}
\begin{table}
	\centering
	\caption{Time complexity of popular interpolation methods to interpolate one point.}
	\label{time_com}
	\renewcommand\arraystretch{1.1}
	\begin{tabular}{p{5cm}|c}
		\toprule[1pt]
		&Time complicity\\
		\hline
		Linear interpolation~\cite{linear_interpolation1}&$O(1)$\\
		Spline interpolation~\cite{cubic_spline}&$O(k)$\\
		Makima interpolation~\cite{makima}&$O(logk)$\\
		PCHIP interpolation~\cite{hermite}&$O(logk)$\\
		RBF interpolation~\cite{rbf}&$O(k^3)$\\
		Kriging interpolation~\cite{kriging}&$O(k^3)$\\
		Ours&$O(k^3)$\\
		\bottomrule[1pt]
	\end{tabular}
\end{table}
To further compare the calculation cost of different interpolation methods, we summarize the time complexity of these interpolation methods in Table~\ref{time_com}~\cite{time_rbf,time_kriging}.
We assume that each interpolation is based on $k$ reference points.
Linear interpolation uses two reference points to establish a linear relationship, so the time complexity is $O(1)$.
Because $k$ is usually a small constant for real-time interpolation, the time complexity of these interpolation methods is similar.
The actual time consumption of various interpolation methods can be seen in Section~\ref{latency_analysis}.

\section{Experimental evaluation}\label{sec:experiment}
This section is devoted to comprehensively analyzing the performance of the proposed interpolation algorithm in terms of accuracy and efficiency.
\subsection{Implementation Details}
To evaluate the efficacy of our proposed interpolation technique, we conduct a comparative analysis with other commonly used interpolation methods using several metrics.
The performance evaluation is carried out on a set of random Bezier curves that vary in length from 1000 to 100,000 data points, with 19 distinct lengths considered.
For each length, we generate ten different random Bezier curves, ranging from the 5th to the 14th order, resulting in a total of 190 random curves.
In addition, to simulate real-world scenarios, we introduce Gaussian noise to the curves, and each curve is evaluated on five different ratios of missing points, such as 10\%, 20\%, 30\%, 40\%, and 50\%.
Additionally, the interpolation results are also assessed both statistically and visually based on the real-scene data collected in the indoor environment.
All the experiments are conducted for five independent runs, and the average results are reported for comparison.
 \subsection{Interpolation Error Analysis}
 \begin{figure}[]
	\centering
	  \includegraphics[width=0.75\textwidth]{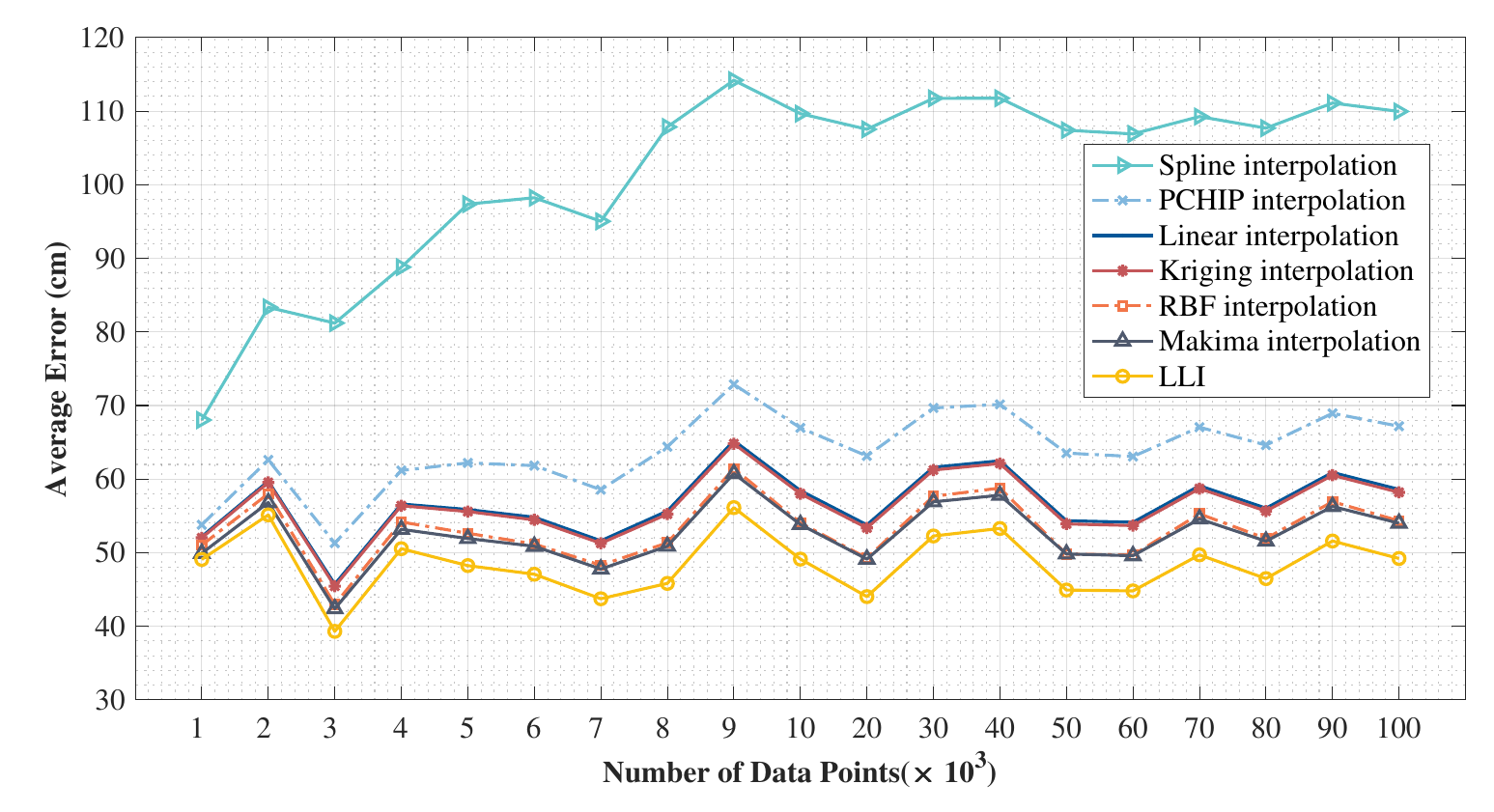}
        \caption{Average positioning errors of various interpolation methods.}
 	\label{average_error}
\end{figure}
 \begin{figure}[]
	\centering
	  \includegraphics[width=0.75\textwidth]{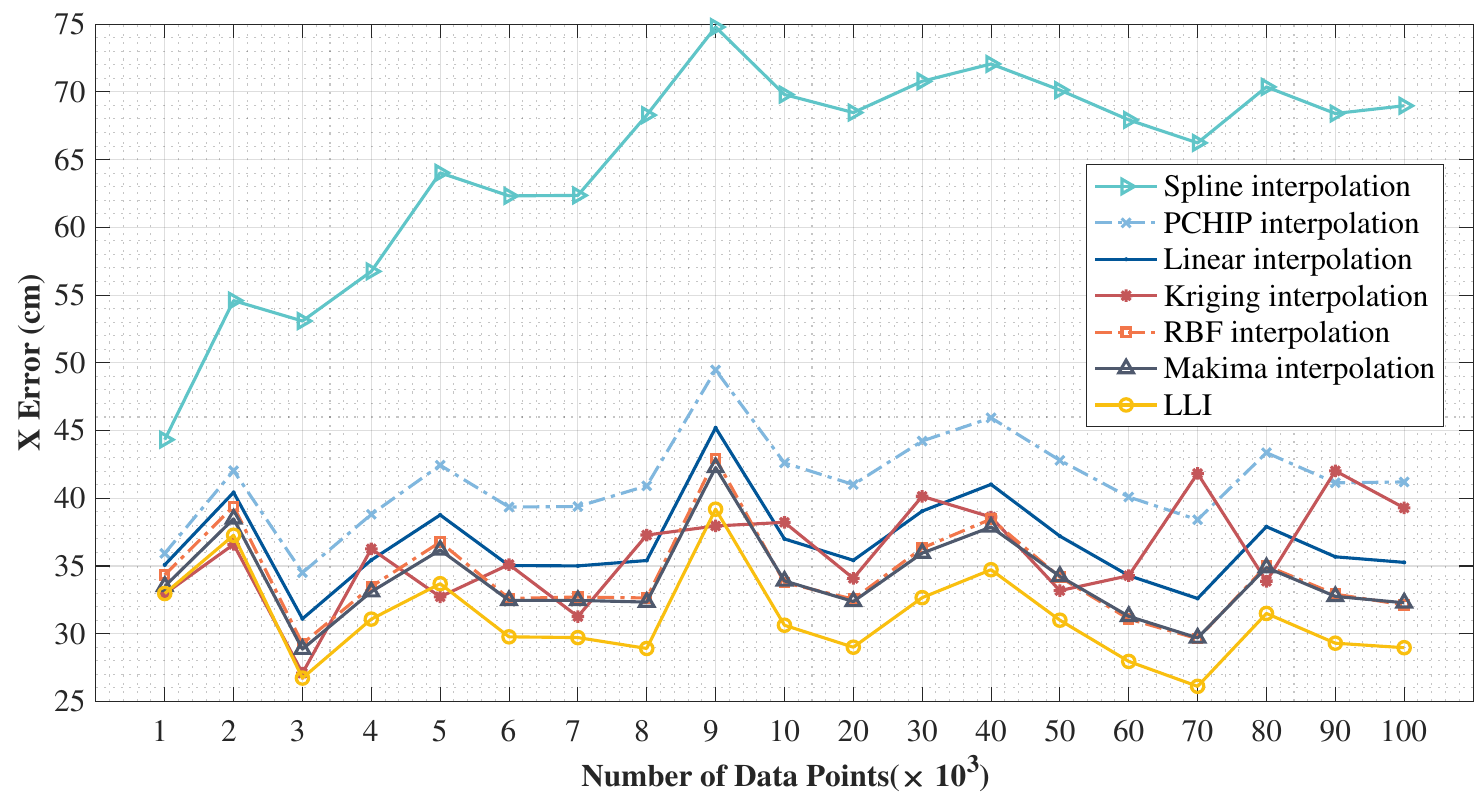}
        \caption{Average positioning X errors of various interpolation methods.}
 	\label{average_x_error}
\end{figure}
  \begin{figure}[]
	\centering
	  \includegraphics[width=0.75\textwidth]{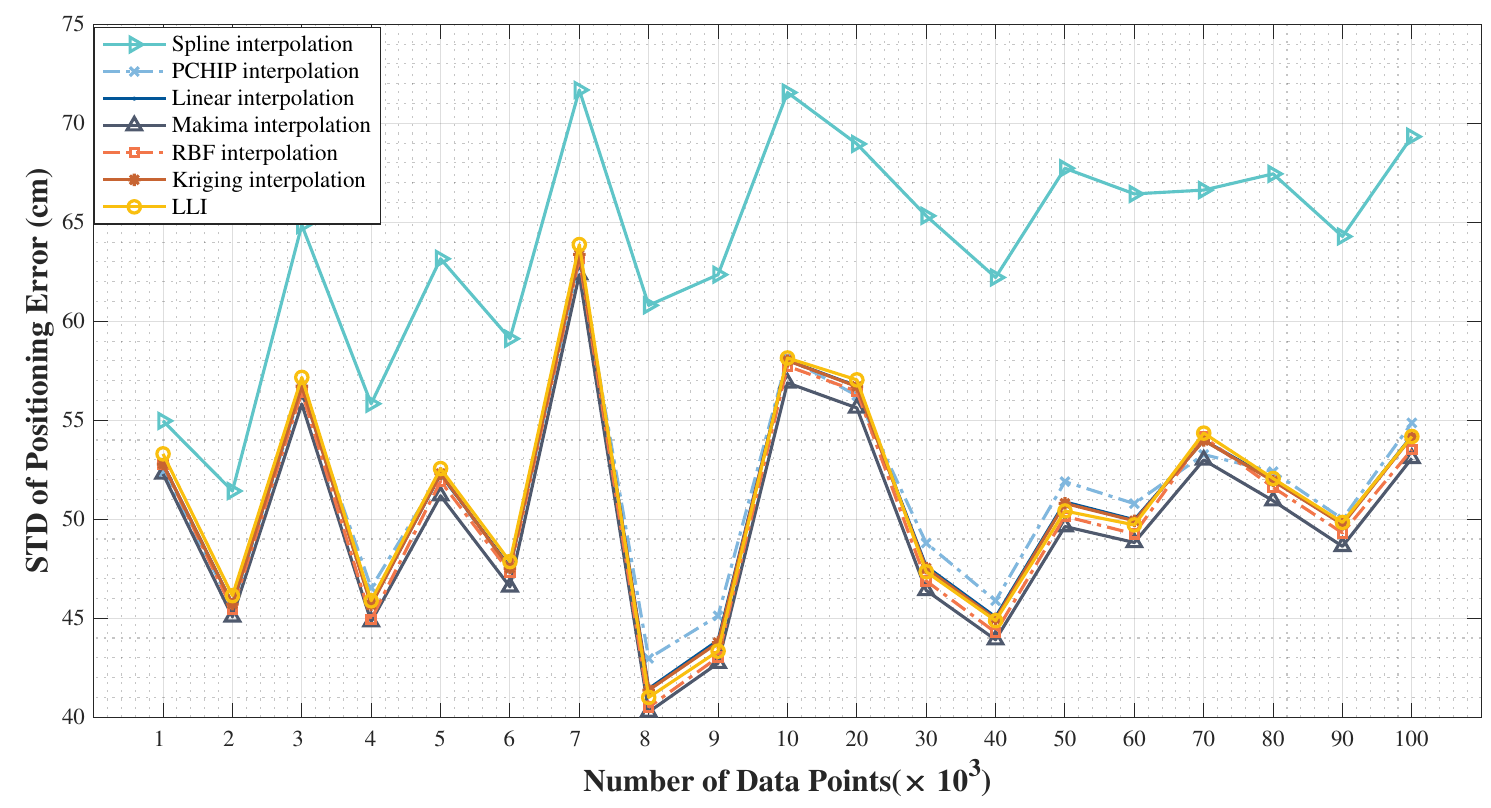}
        \caption{STD of average positioning error.}
 	\label{average_std}
\end{figure}
  In this section, mean square error (MSE) and cumulative distribution functions (CDF) are used as the performance metrics for evaluation.
  Meanwhile,  standard deviation (STD) is also utilized to assess the stability of the performance.
  CDF is defined as follows:
 \begin{equation}\label{cdf}
	\text{CDF}(x)=\frac{1}{N}\sum_{i=1}^{N} \mathbb{I}(\left\| P_i-\hat{P_i} \right\| \geq x ),
\end{equation}
where $P_i$ and $\hat{P_i}$ are the coordinates and interpolated values of the $i$-th point, $N$ is the total number of points to be interpolated, $\left\| . \right\|$ denotes the Euclidian norm, and $\mathbb{I}$ denotes the indicator function.

Figure~\ref{average_error} illustrates the MSE of the interpolated points with various lost-point ratios on a set of random Bezier curves.
Compared to several other interpolation techniques, such as RBF interpolation, Makima interpolation, Kriging interpolation, linear interpolation, PCHIP, and cubic spline interpolation, our proposed method significantly outperforms them regarding interpolation accuracy.
Specifically, when interpolating different numbers of points, our average error remains at the lowest level, suggesting its superior interpolation accuracy.
Therefore, although our method does not introduce complicated mathematical models for interpolation, the proposed method can accurately interpolate missing points by reconstructing the topological manifold of the positioning trajectory.
At the same time, Figure~\ref{average_x_error} presents the average positioning error of abscissa. It can be observed that the fluctuation of average positioning $X$ errors for various methods is more prominent, especially Kriging interpolation.
With 2000, 5000, and 9000 data points, the average positioning $X$ error of Kriging interpolation is slightly lower than that of ours.
However, our average positioning error is always lower than that of Kriging interpolation.
In some cases, Kriging interpolation generates points closer to the abscissa, but the overall Euclidean distance between the interpolated points and the ground truth is farther.
From results in Figure~\ref{average_error} and Figure~\ref{average_x_error}, we can see that ours obtains the state-of-the-art (SOTA) accuracy results among other interpolation methods.

We also evaluate the stability of the interpolation methods by presenting the STD of interpolation errors in Figure~\ref{average_std}.
Remarkably, ours maintains a high level of stability among all methods, with interpolation errors consistently within a narrow range.
Although some methods obtain similar stability with ours (e.g., RBF interpolation, Kriging interpolation, and Makima interpolation), combined with the results in Figure~\ref{average_error}, the interpolation errors of them are generally stable at a relatively high level while the interpolation results of ours can be stabilized within a low error range.
Thus, our results demonstrate that our proposed algorithm achieves the best performance in terms of both accuracy and stability.
\begin{figure}[] 
	\centering
	\includegraphics[width=0.75\textwidth]{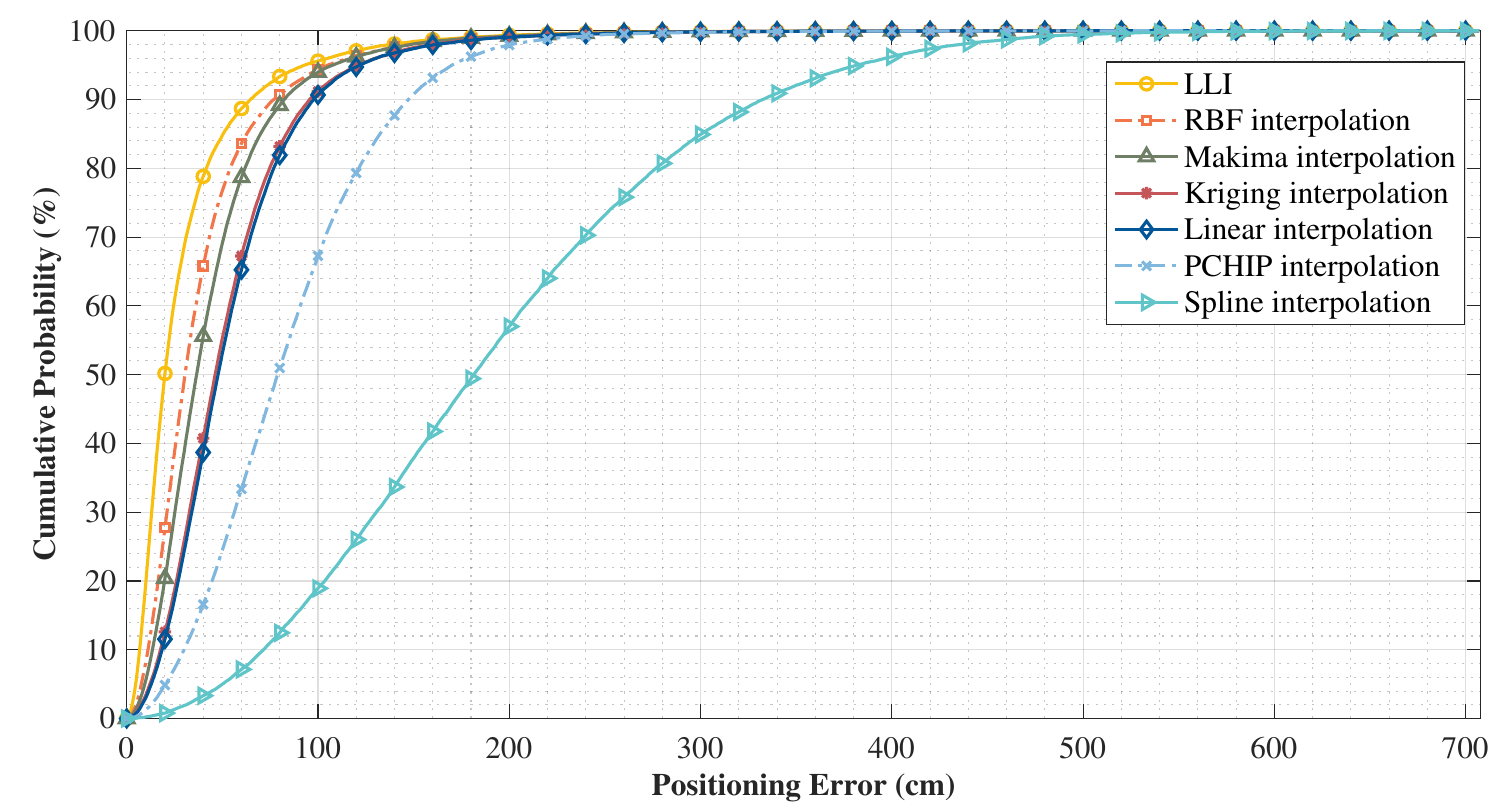}
	\caption{Cumulative error distribution of interpolated points positioning error. }
	\label{CDF}
\end{figure}
\renewcommand\arraystretch{1}
\begin{table}
	\centering
	\caption{The value of $x$(cm) with given CDF($x$) from 50\% to 90\%}
	\label{cdf_table}
	\begin{tabular}{c|ccccc}
		\toprule[1pt]
		The value of CDF	&50\%&60\%&70\%&80\%&90\%\\
		\hline
		Ours&\textbf{17}&\textbf{21}&\textbf{25}&\textbf{31}&\textbf{43}\\
		Linear interpolation~\cite{linear_interpolation1}&45&51&60&69&84\\
		Spline interpolation~\cite{cubic_spline}&181&207&238&275&328\\
		Makima interpolation~\cite{makima}&34&39&45&53&67\\
		PCHIP interpolation~\cite{hermite}&76&87&100&116&138\\
		RBF interpolation~\cite{rbf}&28&32&38&44&56\\
		Kriging interpolation~\cite{kriging}&43&50&58&67&81\\
		\bottomrule[1pt]
	\end{tabular}
\end{table}

 To corroborate further the results shown in Figure~\ref{average_error}, a comparative analysis of the CDF curves of signal interpolation errors is presented in Figure~\ref{CDF}.
It can be observed that the proposed method exhibits a considerably steeper cumulative probability curve than the other methods, indicating that a higher proportion of the interpolated points estimated by ours are within a lower error range.
 To further investigate the accuracy of the methods, Table~\ref{cdf_table} reports the specific error values at various CDF levels. 
 In general, we can observe that the errors of the proposed method are significantly lower than those of other methods at various CDF levels.
 For instance,  at the 90\% CDF level, the error of our method is only 43, which means that 90\% of our interpolation errors are lower than 43.
 In this case, the interpolation error of the proposed method is the lowest among all tested approaches.
 In contrast, for linear and Kriging interpolation, even the error $x$ is already above 43 with just $50\%$ of the CDF($x$).
 Although RBF interpolation and ours exhibit a similar trend in the CDF curves, the overall error of RBF interpolation is much higher than that of ours.
 Specifically, at the 90\% CDF level, the error of RBF interpolation is 56, which is 13 higher than our method's. 
 Consequently, experimental results demonstrate that our method is more accurate and stable than others.
\subsection{Time Consumption}\label{latency_analysis}
In real-time applications, the time consumption of interpolation approaches is a critical performance metric.
To evaluate the efficiency of our proposed method, we conduct a quantitative comparison against other interpolation approaches. 
The experiments use MATLAB on an Inter Core i5-9400 CPU @ 2.90GHZ, 16.00GB RAM computer.

\begin{table}[]
	\centering
	\caption{Time consumption (ms) with different numbers of points.}
	\label{latency_table}
	\begin{tabular}{p{5cm}|ccc }
		\toprule[1pt]
		The number of interpolated points&$3 \times 10^4$& $6 \times 10^4$&$9 \times 10^4$\\
		\hline
		LLI&\textbf{901.41}&\textbf{1662.99}&\textbf{2484.26}\\
		Linear interpolation~\cite{linear_interpolation1}&  $\backslash$ & $\backslash$ & $\backslash$ \\
		Spline interpolation~\cite{cubic_spline}&2810.90&5677.98 &8457.13\\
		Makima interpolation~\cite{makima}&1538.33&3086.28 &4606.12\\
		PCHIP interpolation~\cite{hermite}&1880.86&3774.35&5466.55\\
		RBF interpolation~\cite{rbf}&16760.45&34027.04&51839.18\\
		Kriging interpolation~\cite{kriging}&1958.15&4023.18&5914.37\\
		\bottomrule[1pt]
	\end{tabular}
\end{table} 
Table~\ref{latency_table} presents the time consumption of various interpolation approaches when interpolating different numbers of points (e.g., $3\times10^4$, $6\times10^4$, and $9\times10^4$).
As shown in Table~\ref{latency_table}, the proposed method consistently achieves the lowest time consumption compared to other methods when interpolating different numbers of points. 
While linear interpolation has a negligible time complexity of $O(1)$, it results in a high interpolation error. 
At the same time, Spline interpolation, Makima interpolation, and PCHIP interpolation have time complexities of $O(k)$, $O(\log k)$, and $O(\log k)$, respectively, while others have time complexities of $O(k^3)$~\cite{time_rbf,time_kriging}.
Specifically, when interpolating $9\times10^4$ points,  our algorithm is $70.6\%$ faster than Spline interpolation, $46.1\%$ faster than Makima interpolation, $54.6\%$ faster than PCHIP interpolation, $95.2\%$ faster than RBF interpolation, and $58.0\%$ faster than Kriging interpolation.
This is because our algorithm has no extensive system of equations to solve.
The average time consumption of interpolating a point using our algorithm is less than $0.03$ milliseconds, which is the shortest time except for linear interpolation. 
Therefore, our method is a practical choice for real-time applications such as real-time tracking.
To summarize, the results demonstrate the efficiency and effectiveness of our algorithm compared to other state-of-the-art methods.
\begin{table}
	\centering
	\caption{Time consumption (ms) with different values of $k$.}
	\label{latency_num}
	\begin{tabular}{p{5cm}|ccc }
		\toprule[1pt]
		The value of $k$&5& 10&20 \\
		\hline
		Ours&\textbf{901.41}&\textbf{1101.53}&2116.61\\
		Linear interpolation~\cite{linear_interpolation1}&  $\backslash$ & $\backslash$ &$\backslash$   \\
		Spline interpolation~\cite{cubic_spline}&2810.90&2874.59&2996.44 \\
		Makima interpolation~\cite{makima}&1538.33&1560.23 &\textbf{1619.18} \\
		PCHIP interpolation~\cite{hermite}&1880.86&1993.32&2106.34 \\
		RBF interpolation~\cite{rbf}&16760.45&17888.39&22550.52\\
		Kriging interpolation~\cite{kriging}&1958.15&2465.90&12632.84\\
		\bottomrule[1pt]
	\end{tabular}
\end{table} 

Table~\ref{latency_num} presents the time consumption of different interpolation methods with respect to varying numbers of reference points, $k$.
Generally speaking, the time consumption increases with the increase of $k$.
When $k$ is 5 and 10,  we achieve the lowest time consumption compared to other methods.
Specifically, when $k$ is 5, our time consumption is $67.9\%$ shorter than that of  Spline interpolation, $41.4\%$ shorter than that of  Makima interpolation, $52.1\%$ shorter than that of  PCHIP interpolation, $94.6\%$ shorter than that of   RBF interpolation, and $54.0\%$ shorter than that of   Kriging interpolation.
However, when $k$ is 20, our time consumption increases sharply, which is higher than that of Makima interpolation and PCHIP interpolation.
In this case, our method is only $29.4\%$ faster than Spline interpolation, $90.6\%$ faster than RBF interpolation, and $83.2\%$ faster than Kriging interpolation, but $30.7\%$ slower than Makima interpolation, $0.5\%$ slower than PCHIP interpolation.
This is because the process of our calculation involves matrix inversion.
If the number of reference points increases, our time consumption will increase rapidly. 
Nevertheless, in practice, the number of reference points should not be too large to avoid significant time latency, which will be discussed in Section~\ref{discuss}.
 Lastly, the experimental results demonstrate that we not only achieve the best interpolation accuracy but also incur the least computational burden among widely used techniques.
 Therefore, the proposed method is practical in any indoor positioning system.
\subsection{Real-scene Indoor Positioning Interpolation}
In this section, we present an evaluation of the practical effectiveness of our algorithm in natural indoor positioning scenes, using data collected from IPSs in practice. 
The device used is the Decawave DW1000 UWB transceiver, and a dynamic target is set to simulate the actual application of indoor positioning.
Specifically, the tag is held by a person traveling along an approximately rectangular trajectory with dimensions of about $8m \times 18m$.
\begin{figure*}[]
	\newcommand{\cwidth}{0.28}
	\centering
	\subfloat[LLI]{\includegraphics[width=\cwidth \columnwidth]{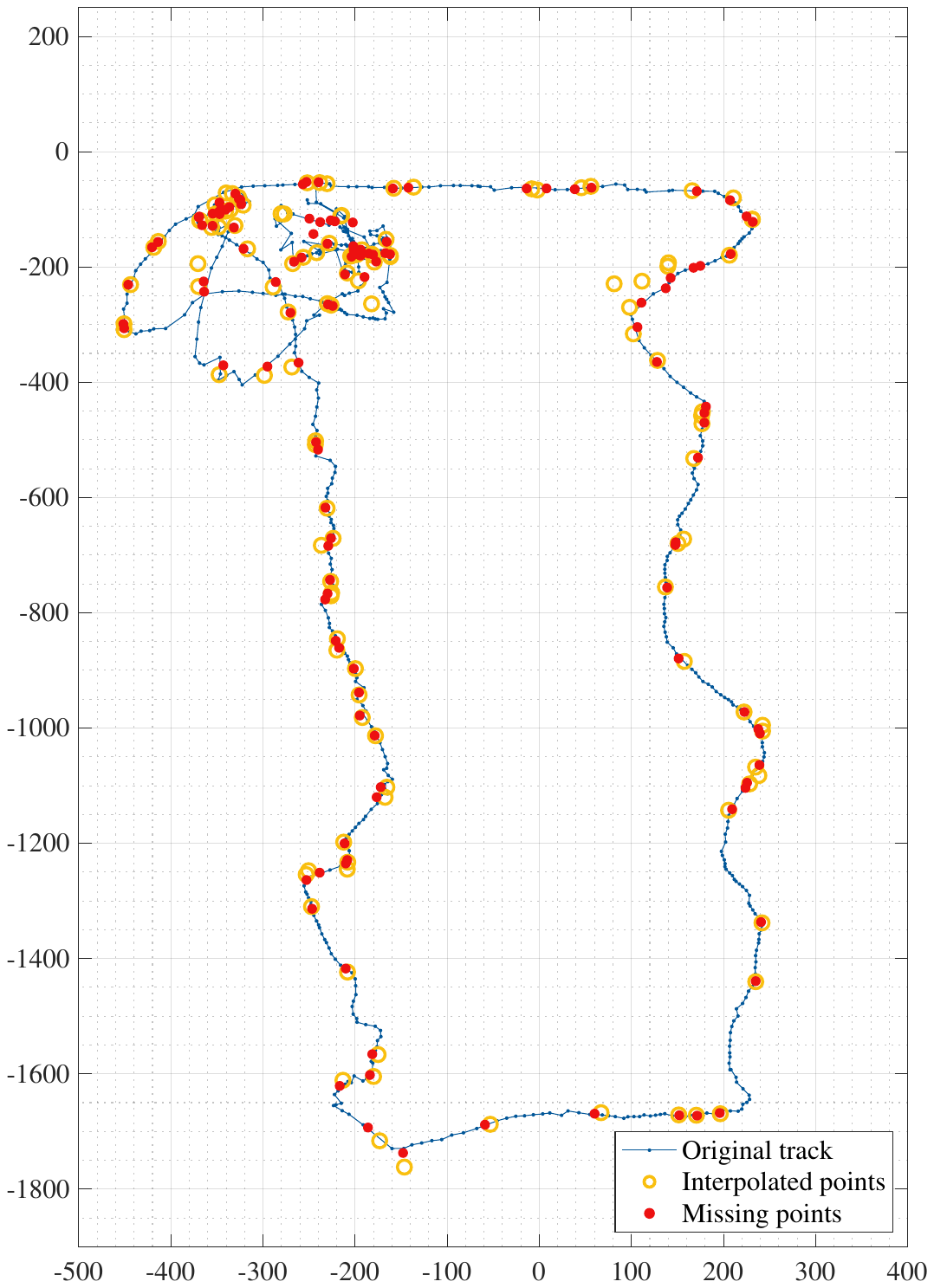} \label{real_LLI}} 
 
	\subfloat[Linear Interpolation]{\includegraphics[width=\cwidth \columnwidth]{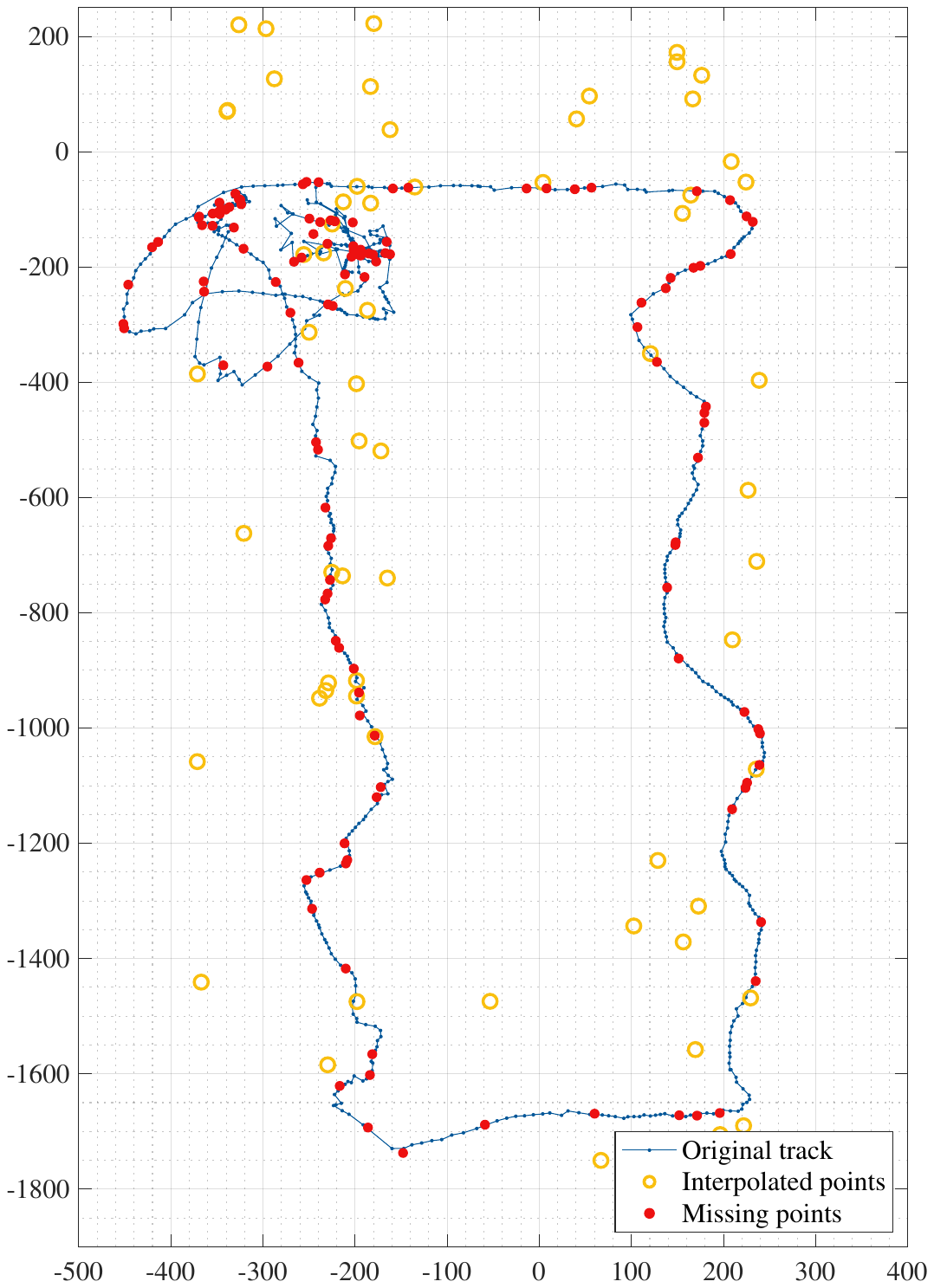} \label{real_linear}} 
	\subfloat[Spline Interpolation] 
	{\includegraphics[width=\cwidth \columnwidth]{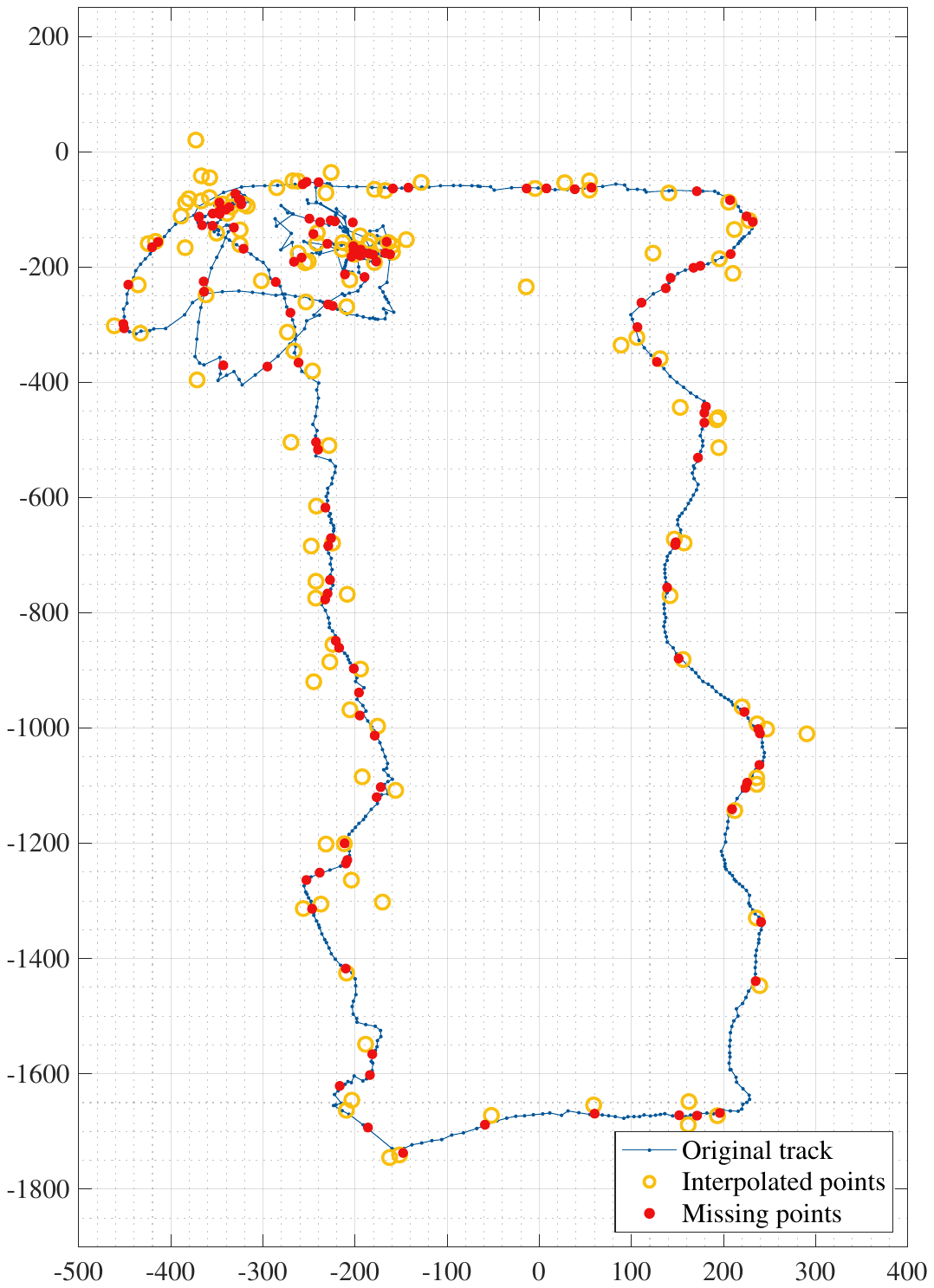}\label{real_spline} } 
	\subfloat[Makima Interpolation]{\includegraphics[width=\cwidth \columnwidth]{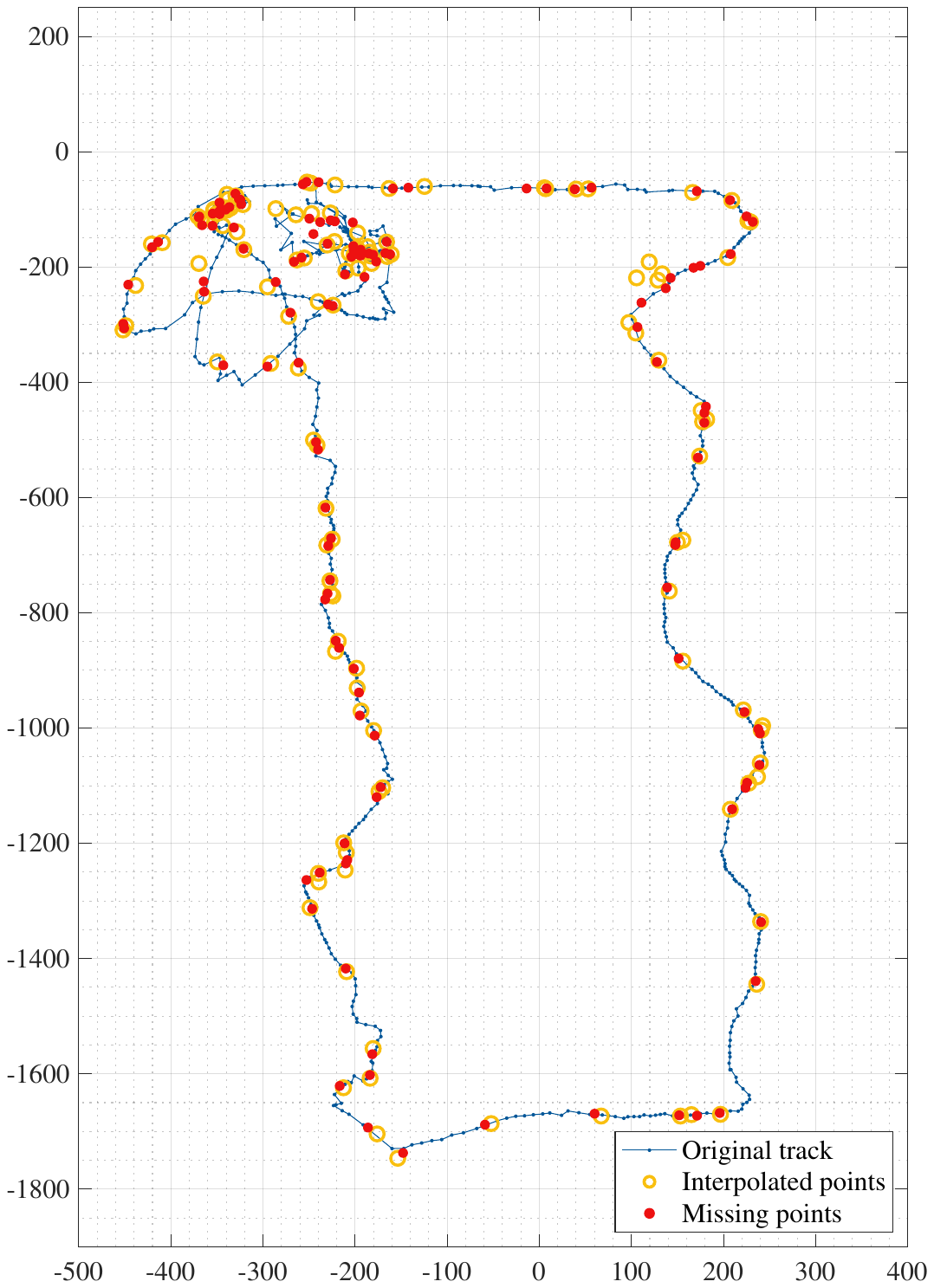}\label{real_makima} }
 
	\subfloat[PCHIP Interpolation]{\includegraphics[width=\cwidth \columnwidth]{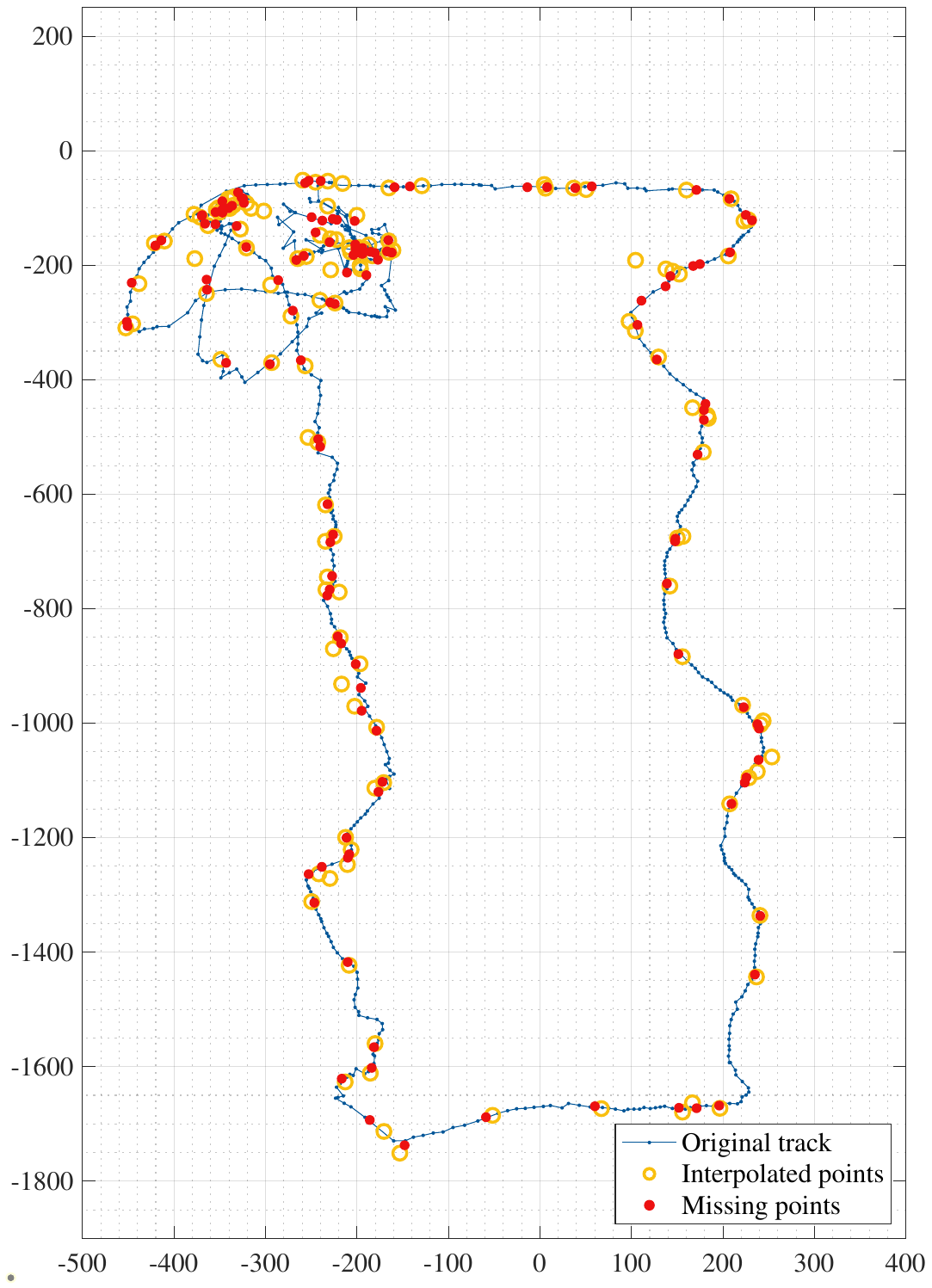}\label{real_pchip}}
	\subfloat[RBF Interpolation]{\includegraphics[width=\cwidth \columnwidth]{pchip_real_scene}\label{real_rbf}}
	\subfloat[Kriging Interpolation]{\includegraphics[width=\cwidth \columnwidth]{pchip_real_scene}\label{real_kriging}}
	\caption{Trajectories with different interpolation methods in real scenarios. We delete 10\% points on the trajectory and interpolate these points with various interpolation methods. } 
	\label{real_eva}  
\end{figure*}
To compare the performance of different interpolation methods, we collected several trajectories and deleted 10\% of the positioning points for interpolation. 
Since there is no ground truth for real-scene data, we use the collected data to estimate the interpolation error, and the statistical results are summarized in Table~\ref{real_scene}.
The statistical results validate that our algorithm outperforms other methods significantly, with an average interpolation error 15.53 smaller than that of cubic spline interpolation, 4.1 smaller than that of Makima interpolation, 4.51 smaller than that of PCHIP interpolation, 10.14 smaller than that of RBF interpolation, and 2.37 smaller than that of Kriging interpolation. 

Meanwhile, the visualization of interpolating results is also presented in Figure~\ref{real_eva}, where each graph depicts the original track, the missing points, and the interpolated points for each interpolation method. 
Specifically, for each graph in Figure~\ref{real_eva}, the blue dots denote the original track, the larger red dots represent the missing points, and the yellow circles indicate the interpolated points.
Ideally, the interpolation points coincide with the missing points; that is, the red dots coincide with the yellow circles.
Our interpolation results are illustrated in Figure~\ref{real_eva}\subref{real_LLI}, where most of the interpolated points are on the trajectory and approximately coincident with the missing points, indicating a good practical effect. 
However, some interpolation points are not too close to the missing points, which may be due to the influence of noises in real scenes.
As can be seen from Figure~\ref{real_eva}\subref{real_linear}, the accuracy of linear interpolation is inferior, because many interpolated points are far from the actual trajectory. 
Cubic spline interpolation obtains similar results to that of linear interpolation.
Lastly, from both statistical and visual results, we demonstrate the practical effectiveness of our method in real-world indoor positioning scenarios, where it achieves superior interpolation performance compared to other methods.
\begin{table}[]
	\centering
	\caption{Average interpolation error in real indoor scenes.}
	\label{real_scene}
	\begin{tabular}{p{5cm}|c}
		\toprule[1pt]
		Interpolation methods&Average interpolation error\\
		\hline
            LLI&\textbf{17.62}\\
		Linear interpolation~\cite{linear_interpolation1}&584.91\\
		Spline interpolation~\cite{cubic_spline}&33.15\\
		Makima interpolation~\cite{makima}&21.72\\
		PCHIP interpolation~\cite{hermite}&22.13\\
		RBF interpolation~\cite{rbf}&27.76\\
		Kriging interpolation~\cite{kriging}&19.99\\
		\bottomrule[1pt]
	\end{tabular}
\end{table}
\subsection{Interpolation within Historical Points Range}\label{within_range}
 can be utilized effortlessly for interpolation within the range of historical data.
In such scenarios, reference points exist before and after the lost point.
Suppose that the positioning point is lost at time $t$ and $t>\lfloor \frac{k}{2} \rfloor +1$.
As $t$ falls within the time range of reference points, we have at least $\lfloor \frac{k}{2} \rfloor +2$ points before and $\lfloor \frac{k}{2} \rfloor $ points after:
\begin{equation}
	P=[\mathbf{p}_{t-\lfloor \frac{k}{2} \rfloor -2},...,\mathbf{p}_{t-1},\mathbf{p}_t,\mathbf{p}_{t+1},...,\mathbf{p}_{t+\lfloor \frac{k}{2} \rfloor}]
\end{equation}
In the process of coefficients reconstruction, we exploit the historical points before and after the lost point to determine the linear relationship:
\begin{equation} 
	\mathbf{p}_{t-1}=\sum_{i=0}^{\lfloor \frac{k}{2} \rfloor} w_{i}*\mathbf{p} _{t-\lfloor \frac{k}{2} \rfloor -2+i}+\sum_{i=\lfloor \frac{k}{2} \rfloor}^{k-1} w_{i}*\mathbf{p} _{t-2+i}
\end{equation}
In this way, the coefficients of abscissa and ordinate are denoted as $W_x$ and $W_y$, respectively.
In the interpolation stage, based on the local topological manifold of the trajectory, we use points $[\mathbf{p}_{t-\lfloor \frac{k}{2} \rfloor -1},...,\mathbf{p}_{t-1}]$ before the lost point and points 
$[\mathbf{p}_{t+1},...,\mathbf{p}_{t+\lfloor \frac{k}{2} \rfloor }]$ after the lost point.
For simplicity,  the weight $W$ is used to construct the lost point as follows:
\begin{equation}
	x=[\mathbf{p}_{t-\lfloor \frac{k}{2} \rfloor -1},...,\mathbf{p}_{t-1},\mathbf{p}_{t+1},...,\mathbf{p}_{t+\lfloor \frac{k}{2} \rfloor }] * W^\intercal
\end{equation}

To evaluate the effectiveness of our method within the historical points range, we examine the performance of different interpolation methods using different lost-point ratios from 20\% to 50\%.
As we can see in Table~\ref{within_acc}, our method gives satisfactory interpolation accuracy with lost-point ratios from 30\% to 50\% consistently.
When the lost-point ratio is 20\%, the interpolation error of the proposed method is 0.26 higher than that of RBF interpolation but lower than other methods.
When the lost-point ratio is above 30\%, our interpolation error becomes the lowest among all algorithms.
Furthermore, the gap in the interpolation error between ours and other methods continues to widen as the lost-point ratio increases.
Specifically, when the lost-point ratio is 30\%, our interpolation error is only 0.62 lower than that of RBF interpolation.
However, when the lost-point ratio is 50\%, our interpolation error is as much as 1.35 lower than that of RBF interpolation.
Thus, these findings suggest that the proposed algorithm outperforms other interpolation methods for interpolating within the historical points range.
Especially as the number of lost points increases, our method demonstrates stronger performance.
\begin{table}[]
	\centering
	\caption{Average interpolation error with different lost-point ratios.}
	\label{within_acc}
	\begin{tabular}{p{5cm}|cccc }
		\toprule[1pt]
		&20\%& 30\%&40\%& 50\% \\
		\hline
		Linear interpolation~\cite{linear_interpolation1}&23.01&22.38&22.32&22.48\\
		Spline interpolation~\cite{cubic_spline}&24.66&23.76&23.57&23.58\\
		Makima interpolation~\cite{makima}&23.46&22.75&22.65&22.76\\
		PCHIP interpolation~\cite{hermite}&23.41&22.69&22.58&22.68\\
		RBF interpolation~\cite{rbf}&\textbf{22.50}&21.98&21.98&22.19\\
		Kriging interpolation~\cite{kriging}&23.30&22.62&22.53&22.66\\
		Ours&22.76&\textbf{21.36}&\textbf{20.98}&\textbf{20.84}\\
		\bottomrule[1pt]
	\end{tabular}
	
\end{table}    

\subsection{Ablation Study}\label{ablation}
\begin{figure}[]
	\centering \includegraphics[width=0.65\textwidth]{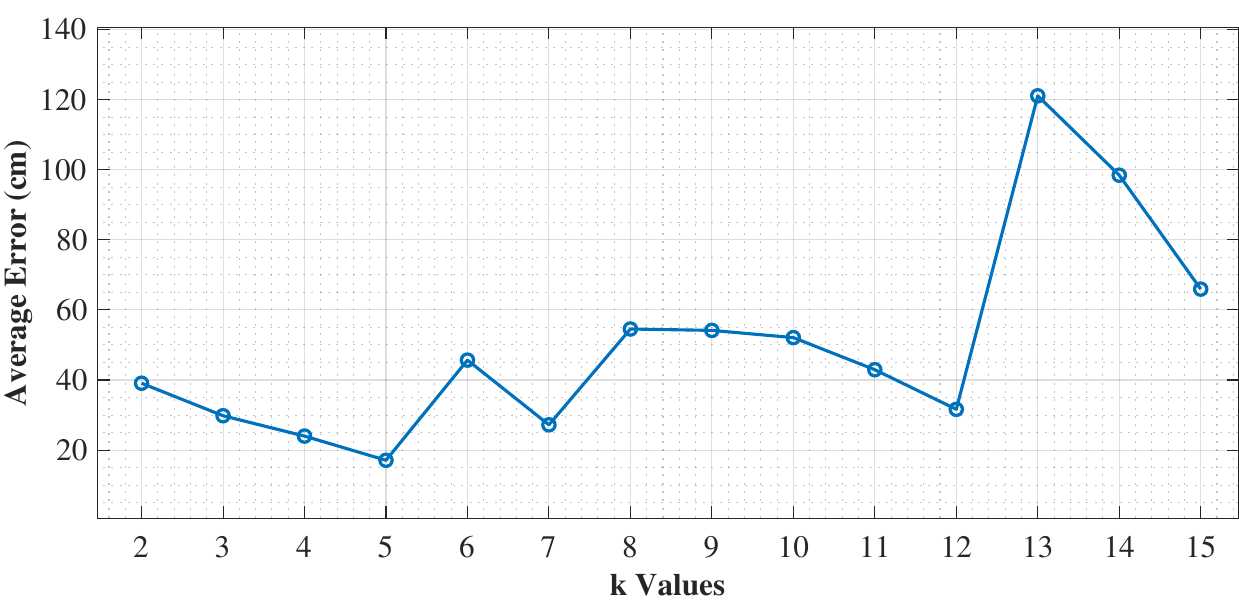}
	\caption{ Interpolation accuracy under different settings of $k$ values.}\label{k_values}
\end{figure}
The parameter $k$ in Equation~\eqref{eq2} closely relates to both the time latency and trajectory smoothness of the interpolation algorithm. 
The value of $k$ determines the number of historical positioning points used to interpolate new points. 
Consequently, smaller values of $k$ reduce the latency of the algorithm, while larger values result in smoother interpolated curves but increase the latency. 
In this section, we investigate the effect of parameter $k$ on interpolation accuracy by varying its value from 2 to 15. 

As depicted in Figure~\ref{k_values}, the interpolation error decreases with the increase of $k$ for $k$ less than 5.
 Specifically, the lowest interpolation error is achieved when $k=5$, with a relatively small time latency.
When $k$ becomes larger, the average interpolation errors fluctuate.
Especially when $k$ is greater than 12, the interpolation error increases sharply, and the latency of interpolation algorithms is also very large.
However, a significant latency is detrimental to applications that require real-time indoor positioning. 
Therefore, in our experiments, we set the value of $k$ to 5.

\section{Discussion}\label{discuss}
The issue of signal interference has long plagued 

In this work, we propose an interpolation approach for indoor positioning tasks.
The proposed method interpolates data points by constructing the local topological manifold within the trajectory of positioning.
In contrast to other methods, our method obviates the necessity for resolving intricate systems of equations, thus conferring substantial practical utility.
Additionally, our algorithm achieves higher interpolation accuracy and enhanced interpolation efficiency among the most widely used and representative methods.
The superior performance is attributed to its consideration of the inherent physical movement characteristics of the object being tracked.
Notably, our method is applicable to a wide range of indoor positioning systems, demonstrating seamless integration capabilities while simultaneously presenting competitive performance.


However, it is imperative to acknowledge certain constraints in our methodology. 
Specifically, at least $k$ reference points are utilized in the interpolation process.
If there are some missing points in the initial $k$ points, interpolation becomes unfeasible.
In fact, the interpolation process remains dormant within the initial $n$ points in practical scenarios, where $n>k$. 

Furthermore, our method necessitates matrix inversion during its computational procedures.
Therefore, setting a significant value for the parameter $k$ can significantly escalate computational expenses of matrix inversion~\cite{wang2020method}, while a minor $k$ influences the smoothness of the interpolated trajectory, which is a tradeoff.
In Section~\ref{latency_analysis}, we have analyzed the influence of the parameter $k$ on the time consumption.
The proposed algorithm is found to be less time-consuming than other interpolation algorithms when $k\leq 10$.
However, when $k=20$, our method experiences a substantial rise in time consumption.  Hence, the selection of the parameter $k$ is pivotal for the interpolation method's overall performance. 
 In our study, we determine the specific value of $k$ by considering the interpolation error (elucidated in Section~\ref{ablation}) and the time consumption (clarified in Section~\ref{latency_analysis}).
 Moreover, $k$ cannot be set too large in real-time indoor positioning endeavors to ensure low latency. 
 Consequently, we choose $k=5$ as a reasonable compromise, balancing interpolation accuracy and temporal efficiency.

Additionally, the proposed method, along with other interpolation methods, confronts a challenge of continuous missing points, which is very common in some environments~\cite{signal_loss}. 
 If the number of consecutive missing points exceeds $k$ or the missing points are densely located, the efficacy of interpolation techniques is compromised.
 The root of this issue lies in the fact that continuous or densely localized data gaps lead to the forfeiture of vital local geometric information. 
Consequently, the interpolation of missing data points becomes contingent solely upon historical positioning data that precedes the gaps, inevitably resulting in substantial time latency and interpolation inaccuracies.
 Future work may, therefore, focus on interpolating accurately and efficiently in these extreme scenarios.
 In such cases, rather than giving equal attention to each historical point, the interpolation should focus more on the latest historical points.
  We may solve this problem by constraining the weighting coefficients of the historical points and compelling the interpolation to give more attention to the latest points, thereby compelling the interpolation process to accord heightened importance to the most recent data points to alleviate the adverse effects of continuous and densely clustered missing data.

\section{Conclusions}
In this work, we proposed a novel and versatile geometric-aware data interpolation algorithm based on manifold learning for indoor positioning.
Instead of constructing complicated mathematical models for interpolation, the proposed algorithm estimates data by exploiting the trajectory's local topological manifold to optimize the interpolation process's computational efficiency while simultaneously enhancing interpolation accuracy.
Through a series of experiments and a meticulous performance analysis, we substantiate the effectiveness of our proposed method by showcasing its superior interpolation accuracy and minimal time consumption when compared to a number of commonly used interpolation methods.
Notably, the superior performance of our algorithm is consistently stable at a relatively high level.
Furthermore, our algorithm exhibits enhanced adaptability across diverse interpolation scenarios, thus broadening its utility. 
Its low time consumption renders it particularly well-suited for real-time applications, meeting the demands of time-sensitive indoor positioning systems.
Future work will involve solving the challenges discussed, such as continuous missing points, which may further expand its application scope and enhance its performance. 

\section*{Data Availability}
Please contact the first author (sryang@smail.nju.edu.cn) to
obtain the Matlab demo codes and data.
\section*{Acknowledgment}
This work was supported in part by the National Natural Science Foundation of China under Grant Nos. (62276127).
\bibliographystyle{elsarticle-num}
\bibliography{reference}
\clearpage
 
 

\end{document}